\documentclass[graybox]{svmult}

\usepackage{mathptmx}       
\usepackage{helvet}         
\usepackage{courier}        
\usepackage{type1cm}        
\usepackage{makeidx}         
\usepackage{graphicx}        
\usepackage{multicol}        
\usepackage[bottom]{footmisc}
\usepackage{stfloats}
\usepackage{multirow}
\usepackage{graphicx}
\usepackage{amsmath}
\usepackage{color}
\usepackage[psamsfonts]{amssymb}
\usepackage{amsxtra}
\usepackage{threeparttable}
\usepackage{footmisc}
\usepackage{url}
\usepackage{booktabs}
\makeindex             

\begin{document}

\title*{Generating Sentiment-Preserving Fake Online Reviews Using Neural Language Models and \\ Their Human- and Machine-based Detection}
\titlerunning{Generation and Detection of Fake Online Reviews}

\author{
David Ifeoluwa Adelani$^1$,
Haotian Mai$^2$,
Fuming Fang$^3$,
Huy H. Nguyen$^{3,4}$, 
Junichi Yamagishi$^{3,4}$, and
Isao Echizen$^{3,4}$ \\
$^1$Spoken Language Systems (LSV), Saarland Informatics Campus, Germany \\
$^2$University of Southern California, USA\\
$^3$National Institute of Informatics, Tokyo, Japan \\
$^4$SOKENDAI, Kanagawa, Japan
}
\authorrunning{Adelani et. al.}

%
\maketitle

\abstract{Advanced neural language models (NLMs) are widely used in sequence generation tasks because they are able to produce fluent and meaningful sentences. They can also be used to generate fake reviews, which can then be used to attack online review systems and influence the buying decisions of online shoppers. 
To perform such attacks, it is necessary for experts to train a tailored LM for a specific topic. In this work, we show that a low-skilled threat model can be built just by combining publicly available LMs and show that the produced fake reviews can fool both humans and machines.
In particular, we use the GPT-2 NLM to generate a large number of high-quality reviews based on a review with the desired sentiment and then using a BERT based text classifier (with accuracy of 96\%) to filter out reviews with undesired sentiments. Because none of the words in the review are modified, fluent samples like the training data can be generated from the learned distribution. A subjective evaluation with 80 participants demonstrated that this simple method can produce reviews that are as fluent as those written by people. It also showed that the participants tended to distinguish fake reviews randomly. Three countermeasures, Grover, GLTR, and OpenAI GPT-2 detector, were found to be difficult to accurately detect fake review. 
}

\vspace{-5mm}
\section{Introduction}
\vspace{-5mm}
\label{sec:introduction}
Neural text generation is one of most active research areas in deep learning. It involves building a neural network based language model (known as {\it neural language model} (NLM)~\cite{bengio2003neural}) given a set of training text token sequences and then using the learned model to produce texts similar to the training data. With the development of deep learning algorithms, neural text generation has become an indispensable technique in the natural language processing field as it can generate more fluent and semantically meaningful text than conventional methods~\cite{ntg_survey}. Its application mainly includes machine translation~\cite{BahdanauCB14}, image captioning~\cite{Vinyals_show_and_tell}, text summarization~\cite{see-etal-2017-get}, dialogue generation~\cite{serban+al-2016-aaai}, and speech recognition~\cite{sundermeyer2012lstm}.

High-performance NLMs can also be used to generate fake reviews, fake comments, and fake news. The generated fake reviews, fake comments, or fake news can then be used to attack online systems or fool human readers. For example, a review system can be flooded with positive reviews to increase a company's profit~\cite{Luca2016FakeIT} or with negative reviews to reduce a competitor's profit, and fake comments/news can be posted on social websites for political benefits. Previous work~\cite{yao2017automated,juuti2018stay} demonstrated the feasibility of fake review attacks. However, 
because these methods apply basic language models (LMs) and take some keywords or meta information as input,
post-processing was needed to adjust the contents to match the desired topic. 
This means that professional experts are needed to train a high performance LM and design an additional language processing method.
It is, therefore, interesting and is also more risky if a threat model with low-skilled but high-performance LM can be easily built.
We hypothesize that this could be feasible since there are many state-of-the-art pre-trained high-performance LMs shared on the Internet for reproducing the experimental results in the literature. With such publicly available LMs, it is possible for non-experts to build powerful threat models.
In this work, we show an example of such threat models and evaluate human- and machine-based detection.

\begin{figure}[tb]
    \centering
    \includegraphics[width=0.6\textwidth]{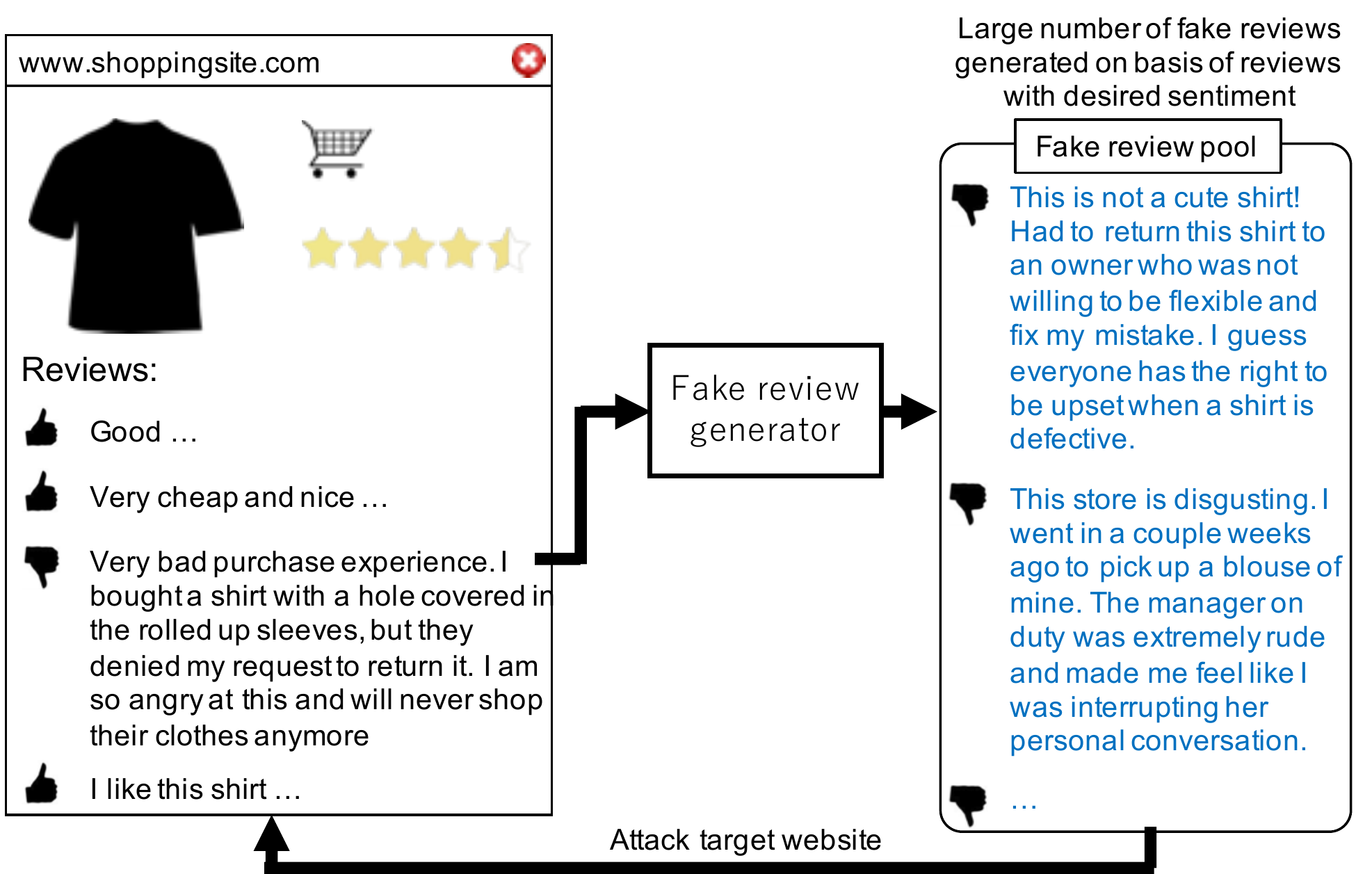}
    \vspace{-2mm}
    \caption{Threat model proposed in this work. A review with the desired sentiment (positive or negative here) is taken from the target shopping website automatically and input to a fake review generator to produce a large number of fake reviews with the same sentiment.}
    \label{fig:threat_model}
    \vspace{-5mm}
\end{figure}

Figure~\ref{fig:threat_model} shows the threat model proposed in our investigation. We suppose that an attacker is able to access reviews (or comments) on a website (e.g., a shopping website) and use a method to automatically identify reviews with a desired sentiment (i.e., positive or negative in this work). We also suppose that the attacker can access a large database containing real reviews (written by people) to train an LM for automatic text generation. The attacker then inputs the identified reviews to the LM to generate a large number of fake reviews. The generated reviews that have the same sentiment as the original review are selected to a fake review pool. Since the fake reviews are generated on the basis of an original review, the context of the original review (e.g., an Italian restaurant) should be implicitly embedded in them. Finally, the attacker submits the selected fake reviews to the site to increase or decrease the rating of a product, service, etc.

To generate sentiment-preserved fake reviews, we use a pre-trained GPT-2 NLM~\cite{radford2019language}, which is able to generate length variable, fluent, meaningful sentences, to generate reviews and then use a fine-tuned text classifier based on BERT~\cite{devlin2018bert} to filter out undesired-sentiment reviews. Since GPT-2 training data differs from the data used in our experiment (i.e., Amazon reviews~\cite{he2016ups} and Yelp reviews~\cite{zhang2015character}), it may generate reviews with irrelevant topic. We solved this problem by adapting the original GPT-2 model to the two databases we used. Subjective evaluation with 80 participants demonstrated that the fake reviews generated by our method had the same fluency as those written by people. It also demonstrated that it was difficult for the participants to identify fake reviews given that they tended to randomly identify fake reviews as the one most likely to be the real review. 
Automatic detection with three countermeasures, Grover~\cite{zellers2019neuralfakenews}, GLTR for detecting text generated by an LM~\cite{Gehrmann2019GLTRSD}, and OpenAI GPT-2 detector~\cite{solaiman2019release}, was also investigated.
The results reveal that automatic detection has better performance than humans, but, the accuracy of detection of the fake reviews is far from perfect and has significant room for improvements.

\vspace{-7mm}
\section{Related Work}
\vspace{-5mm}
\label{sec:related_work}
The most common attack on online review systems is a crowdturfing attack~\cite{lee2014dark,levchenko2011dirty} whereby a bad actor recruits a group of workers to write fake reviews based on a specified topic for a specified context and then submits them to the target website. Since this method has an economic cost, it is typically limited to large-scale attacks. Automated crowdturfing, in which machine learning algorithms are used to generate fake review, is a less expensive and more efficient way to attack online review systems.

Yao et al.~\cite{yao2017automated} proposed such an attack method. Their idea is to first generate an initial fake review based on a given keyword using a long short-term memory (LSTM)-based LM. Because the initial fake review is stochastically sampled from a learned distribution, it may be irrelevant to the desired context. Then specific nouns in the fake review are replaced with ones that better fit the desired context. Juuti et al.~\cite{juuti2018stay} proposed a similar method for generating fake reviews that further requires additional meta information such as shop name, location, rating, and etc.

Our method differs from these methods in that we use a whole review as the seed for generating a large number of fake reviews without using additional information or additional processing and then filter out the ones without the desired sentiment. Our method is thus more straightforward. We do not modify the generated reviews, so their fluency is close to that of the training samples. Since the LM used is adapted from a pre-trained model, our method can be easily implemented even by low-skill attackers.

In addition, adversarial text examples can also be used for attacking online review systems~\cite{liang2017deep,ebrahimi2017hotflip}. The aim is to deceive text classifiers, not people, by adding small perturbations to the input. Unlike this type of method, fake reviews generated by our method are aimed at changing overall user impressions.

\vspace{-7mm}
\section{Fake review generation}
\vspace{-5mm}
The most important part of the proposed method for generating sentiment-preserving fake reviews is the \textit{GPT-2} text generation model~\cite{radford2019language}. Details of our method are as below.

\subsection{GPT-2 Model}
\vspace{-5mm}
The task of an LM is to estimate the probability distribution of a text corpus or to estimate the probability of the next token conditioned on the context tokens. Given a sequence of tokens $\textbf{x} = (x_1,..., x_T)$, the probability of the sequence can be factorized as
\vspace{-1mm}
\begin{equation}
	P(\bold{x}) =  \prod_{t=1}^T P(x_t | x_1,...,x_{t-1}).
\label{eq:lm}
\end{equation}
\vspace{-1mm}

This probability is approximated by learning the conditional probability of each token given a fixed number of $k$-context tokens by using a neural network with parameters $\Theta$. The tokens used for training can be of different granularities such as word~\cite{Bengio2003}, character~\cite{Kim2016}, sub-word unit~\cite{sennrich-etal-2016-neural}, or hybrid word-character~\cite{VerwimpPhW17}. The objective function of the LM is to maximize the sum of the logs of the conditional probabilities over a sequence of tokens: 
\begin{equation}
\vspace{-1mm}
\Theta^* = \underset{\Theta}{\mathrm{argmax}}
 \  \sum_{t=1}^T log P(x_t | x_{t-k},...,x_{t-1}; \Theta).
\label{eq:lm}
\vspace{-1mm}
\end{equation}
The neural network parameters $\Theta$ can be learned using various architectures such as a feed-forward neural network~\cite{Bengio2003}, a recurrent neural network (RNN) such as a vanilla RNN~\cite{Mikolov2010, Shen2017EstimationOG}, an LSTM~\cite{sundermeyer12_lstm} and its variants~\cite{mlstm}, and the transformer~\cite{transformer, gpt} architectures. A GPT-2 model based on the transformer architecture has the lowest perplexity on various language modeling datasets and it generates high-quality fluent texts.

The GPT-2 model was trained on a large unlabeled dataset --- 8 million webtexts obtained by scraping all outbound links (about 45 million) from Reddit, resulting in about 40 GB of text. This LM is easily \textit{generalizable} to a corpus for domains that differ from that of the original training data. For instance, the GPT-2 LM attained state-of-the-art lower perplexity on seven out of eight tested datasets in a zero-shot setting. In addition, generative pre-trained models such as GPT-2 are \textit{transferable} to many natural language understanding tasks such as document classification, question answering, and textual entailment through discriminative fine-tuning of the models within a few epochs. Moreover, the GPT-2 LM can be adapted to a new domain by fine-tuning the model on a corpus in that domain, e.g., online reviews.

There are four different GPT-2 models in terms of size. We used the smallest one (117 million parameters, $\Theta$)\footnote{\url{https://github.com/openai/gpt-2}}. As of now, they have released only the smaller models --- 117M and 345M --- to prevent the malicious use of their larger models. Even with the smallest one, we were able to generate realistic reviews.

\vspace{-7mm}
\subsection{Sentiment-Preserving Fake Review Generation}
\vspace{-5mm}

\begin{figure}[tb]
 \centering
\includegraphics[width=0.7\columnwidth]{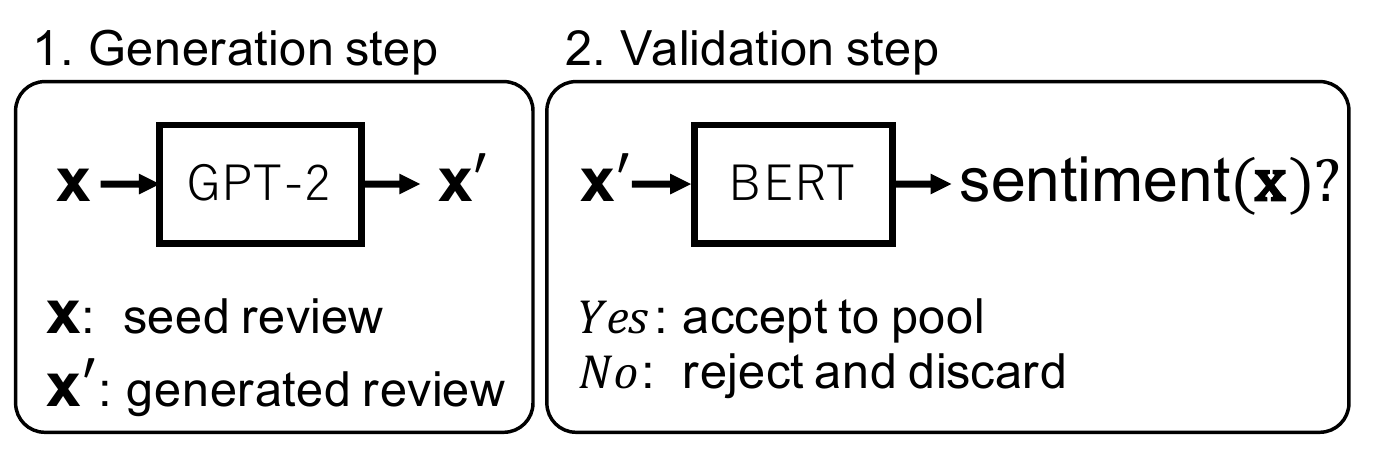}
\vspace{-4mm}
\caption{Fake review generation procedure}
\label{Fig:fake_review_attack}
\vspace{-3mm}
\end{figure}

As shown in Figure~\ref{Fig:fake_review_attack}, we use a two-step approach to generating sentiment-preserved reviews: generation and validation. In the generation step, the attacker provides an original review ${\bold x}$ with a given sentiment as the seed text to the GPT-2 LM, which then generates a different  review ${\bold x}'$ based on ${\bold x}$. We refer to ${\bold x}'$ as a \textit{fake review}; it differs from ${\bold x}$ in its literal representation. There is no strict guarantee that the original review and the fake review have the same context because ${\bold x}'$ is sampled from the probability distribution represented by the model while the context information may be implicitly embedded in ${\bold x}'$ to some degree. Therefore, part of ${\bold x}'$ can be thought of as a continuation or paraphrase of ${\bold x}$.

Validation step aims to filter out the generated reviews with undesired sentiment.
In this step, the attacker determines whether ${\bold x}'$ has the same sentiment as ${\bold x}$ by using the BERT text classifier~\cite{devlin2018bert}, which is similar to the GPT-2 in that it is also based on the transformer, but it further takes into account bidirectional context information. We assume that the attacker has access to such a classifier and uses it to quickly check the generated reviews for their sentiment. If the sentiment of ${\bold x}'$ is the same as that of ${\bold x}$, it is added to the fake review pool. Otherwise it is discarded.

\vspace{-7mm}
\subsection{Fine-tuning Language Model on Review Data}
\vspace{-5mm}
One major advantage of LMs like GPT-2 is that they are very easy to adapt (i.e., fine-tune) to a new dataset or domain. During fine-tuning, the model is first initialized before training with the pre-trained parameters rather than random weights. Fine-tuning takes less time than training a high-capacity LM from scratch with millions of web documents. Furthermore, text classification and other natural language understanding tasks benefit from pre-training the model on a large amount of unlabeled text. It has been shown that fine-tuning using labeled data after initializing the model with pre-trained parameters improves accuracy for downstream tasks~\cite{devlin2018bert}. Therefore, we fine-tuned both the GPT-2 LM and the BERT classifier. We used Amazon and Yelp review databases containing both positive and negative reviews written in English. Following the approach of Yang et al.~\cite{zhang2015character}, we divided the reviews in each database into training and test datasets, as shown in Table~\ref{Tab:data_stat}. The model was fine-tuned on each training dataset, and evaluation was performed on the respective test dataset.

\begin{table}[tb]
\begin{center}
  \caption{Statistics for Amazon and Yelp review databases used for fake review generation.} 
  \scalebox{0.9}{
  \begin{tabular}{p{40mm}p{20mm}r}
  \toprule
    &  Amazon &  Yelp\\     \midrule
    Total number of reviews & 4 million  &  598, 000\\
    Number of training examples & 3.6 million & 560, 000\\
    Number of test examples & 400, 000 & 38, 000\\
    Number of class labels & 2  & 2  \\
    \bottomrule
    \end{tabular}
    }
    \label{Tab:data_stat}
    \end{center}
    \vspace{-4mm}
\end{table}

\begin{table}[tb]
  \begin{center}
  \footnotesize
  \caption{Example reviews generated using pre-trained GPT-2 LM.}
  \scalebox{0.9}{
  \begin{tabular}{p{40mm}p{75mm} }
    \toprule
    \textbf{Method} & \textbf{Seed (from Amazon)/generated review }   \\ 
    \midrule
     
    \multirow{3}{*}[0.5em]{Original Review (SEED)} &  I currently live in europe, and this is the book I recommend for my visitors. It covers many  countries, colour pictures, and is a nice starter for before you go, and once you are there.
    \\ \addlinespace[1.0em]
    \multirow{3}{*}{Pre-trained GPT-2  fake review} & Just as I recommend before you go. And there are lots more things to read. What are your favourite books of the day?  This is my take on the day before a work trip to
    \\ \addlinespace[1.0em]
    \multirow{3}{*}{Fine-tuned GPT-2 fake review} & Great for kids too. Recommended for all young  people as the pictures are good (my kid's are  11) favourite books of the day?  This is my take on the day before a work trip
    \\
    \bottomrule
  \end{tabular}
  }
  \label{tab:issues_pretrainedGpt}
  \end{center}
  \vspace{-7mm}
\end{table}

As of now, the authors of GPT-2 have not released the training code, but we found a reliable source code\footnote{\url{https://github.com/nshepperd/gpt-2}} on GitHub for training the GPT-2 model, which is the implementation we used to fine-tune the pre-trained model on the review databases. We fine-tuned the GPT-2 by concatenating all reviews with a \textit{newline} symbol into a giant text file; we did not distinguish between positive and negative reviews during fine-tuning. We fine-tuned the 117M GPT-2 model on the Amazon training set for two weeks (485K epochs) and on the Yelp training set for five days (190K epochs) by using the default hyper-parameters. We stopped the training when the validation error was no longer decreasing. We found that the pre-trained GPT-2 LM sometimes produced texts that were not review-like, as shown in Table~\ref{tab:issues_pretrainedGpt}. Nevertheless, after fine-tuning, the generated texts were review-like.

Similarly, we fine-tuned the BERT text classifier on the Amazon and Yelp training sets for three epochs to classify reviews as positive or negative. We achieved $96.2$\% accuracy on the original Amazon test dataset and $96.0$\% accuracy on the original Yelp test dataset. Fine-tuning BERT took only a few hours, and the performance was better than that reported for the character-level CNN~\cite{zhang2015character} ($94.49$\% for the Amazon test dataset; $94.11$ \% for the Yelp test dataset).

\vspace{-7mm}
\subsection{Explicit sentiment modeling}
\vspace{-5mm}
In addition to the above basic attack method, which simply fine-tunes the pre-trained GPT-2 LM, we further propose a ``skill-up'' method in which an LM is explicitly conditioned by a specified sentiment. This method requires a natural language processing expert to train a tailored LM.

Radford et al.~\cite{RadfordJS17} reported that a sentiment neuron can be learned by using a single-layer multiplicative LSTM (mLSTM)~\cite{mlstm}. The sentiment neuron can be found by manually visualizing the distribution of output values of hidden units, and a unit for which the output values can be categorized into two groups across multiple sentiment databases can be considered as a sentiment neuron. It has reported that mLSTM outperforms LSTM because it allows each possible input to have different recurrent transition functions~\cite{mlstm}, so fake review generation based on mLSTM is better than that based on LSTM~\cite{yao2017automated}. By replacing the output values of the sentiment neuron with $+1$ (positive) or $-1$ (negative), we can explicitly force the output to be conditioned by a specified sentiment~\cite{RadfordJS17}. We refer to this method as ``sentiment modeling". Our implementation is based on that of Puri et al.~\cite{Puri2018LargeSL}\footnote{https://github.com/NVIDIA/sentiment-discovery}, which had 4,096 units.

\vspace{-7mm}
\section{Experiment}
\vspace{-5mm}
\subsection{Measurements and Setup}
\vspace{-5mm}
We measured the effectiveness of the proposed method for generating sentiment-preserving fake reviews in three ways. 1) The \textit{sentiment-preserving rate} was used for evaluating whether the sentiment of the original review was preserved, with the BERT text classifier used for sentiment prediction. It was defined as the ratio of number of sentiment correctly preserved fake reviews to number of total fake reviews. Note that all generated reviews (without filtering) were used. 2) \textit{Subjective evaluation} was used for evaluating the fluency of the generated reviews and how well people could distinguish between the real reviews and the fake ones. 3) The \textit{detection rate} was used for evaluating how well machine-based detection methods could identify fake reviews.

Four types of LMs were investigated: a pre-trained GPT-2 LM, a fine-tuned GPT-2 LM, an mLSTM LM, and a sentiment modeling. Considering the high computational cost, we randomly selected 1,000 reviews from each test dataset for use as seed texts under the assumption that most of the reviews were written by a person. For each LM, we then generated 20 different fake reviews based on each real review. In total, there were 20,000 fake reviews per LM per dataset. The generated reviews contained from 1 to 165 words, with an average of 94 words. Training of the LMs and review generation were performed on a machine with a Tesla P100 GPU.

For the subjective evaluation, we first asked 80 volunteers (39 native and 41 non-native English speakers) to evaluate the fluency of reviews. Fifty real reviews (200 $-$ 300 characters) were randomly selected (half were positive and half were negative) from each test dataset, and fake reviews were generated on the basis of those reviews. We used the real reviews and the fake reviews with a sentiment most closely matching the associated real review for fluency evaluation. The evaluation was done using a 5-point Likert mean opinion score (MOS) scale, with 5 being the most fluent. We then asked them to select from four reviews the one they thought was the most likely real review, where the four reviews contain a real review and three fake reviews. The average correct selection rate was used as the metric. To facilitate evaluation, the reviews were shortened to only the first three sentences. The evaluations were performed on a web interface\footnote{An image of the interface is available at \url{https://nii-yamagishilab.github.io/fakereview_interface/}} with the real and fake reviews listed in random order. The participants evaluated a minimum of 10 and a maximum of 100 random reviews. Most of the participants evaluated only ten reviews. We obtained 1025 data points for fluency and real/fake selection evaluation, respectively.

For machine-based fake review detection, we used the fine-tuned GPT-2 LM as the text generation model and we used the Grover~\cite{zellers2019neuralfakenews}, the GLTR~\cite{Gehrmann2019GLTRSD}, and the OpenAI GPT-2 detector~\cite{solaiman2019release} as countermeasures. The Grover is based on a neural network and it can defend against fake news generated by an NLM such as the GPT-2 LM. Its reported detection accuracy is 92\%. The GLTR does not directly judge whether text is real or fake. Rather it helps a person to distinguish real from fake text by reporting how likely a word in the text was machine generated.
This tool assigns one out of four labels for each word. These labels could be the top-10, top-100, or top-1000 most frequent words used by a machine, or the least frequently used words. We concatenate numbers of each of assigned labels as a four-dimensional feature vector and then input it to a regression model to tell fake and real reviews apart.
The OpenAI GPT-2 detector is based on RoBerta~\cite{liu2019roberta} and we call it as GPT-2PD.
We further fused these detectors using logistic regression at the score level.
Equal error rate (EER) was used for measuring performance of these detectors.

\vspace{-7mm}
\subsection{Sentiment-preserving fake review analysis}
\vspace{-5mm}

\begin{table}[tb]
    \caption{Rate (in \%) and standard error of fake reviews preserving sentiment of original review. 
       }
    \label{Tab:accuracy}
    \footnotesize
    \centering
   \begin{tabular}{p{30mm}p{20mm}r}
   \toprule
    LM & Amazon  &  Yelp\\
    \midrule
    Pretrained GPT-2            & $62.1 \pm 0.9$   &  $64.3 \pm 1.4$  \\
    Fine-tuned GPT-2     & $67.0 \pm 1.4$   &  $67.7 \pm 1.2$   \\ 
    mLSTM  & $63.2 \pm 0.7$   &  $71.0 \pm 1.3$  \\
    Sentiment modeling   & $70.7 \pm 1.3$  &  $70.1 \pm 1.2$   \\
    \bottomrule
    \end{tabular}
    
    \vspace{-5mm}
\end{table}

As shown in Table~\ref{Tab:accuracy}, the fine-tuned GPT-2 model was better at preserving the sentiment of the original review than the pre-trained GPT-2 model for both databases. This means that a large number of fake reviews can be efficiently generated with a desired sentiment by just fine-tuning an LM. The sentiment modeling method had the highest rate for the Amazon database. This was because explicitly modeling sentiment benefits from the additional sentiment information given before the fake reviews are generated. This indicates that explicitly modeling sentiment could be a more efficient way to generate desired sentiment reviews. For the Yelp reviews, fine-tuned GPT-2 was also clearly better than the pretrained GPT-2 and the mLSTM had the highest rate. Further analysis revealed that the mLSTM model performs very well only for food and restaurant reviews but it did not generalize well to other domains or it generates reviews completely outside the context of the original review. As shown in Table~\ref{tab:fake_review_Example}, the fake review generated by the mLSTM model review talks about a restaurant experience while the original review was about a shirt purchase. This suggests that we need to further explicitly preserve context. (We leave this for future work.) 

\begin{table*}[tb]
  \footnotesize
  \centering
    \caption{Examples of fake reviews generated by four models using original review as the seed. Bold font indicates words expressing sentiment.}
  \label{tab:fake_review_Example}
   \scalebox{0.9}{
  \begin{tabular}{p{32mm}p{95mm}}
    \toprule
    \textbf{Seed / Model} & \textbf{Yelp} \\ 
    \midrule
     
    \multirow{3}{*}[0.2em]{Original Review (SEED)} &
    \textbf{Very bad} purchase experience. I bought a shirt with a hole covered in the rolled up sleeves, but  they denied my request to return  it. I am so \textbf{angery} at this and will \textbf{never shop} their chothes anymore.\\
    
    \addlinespace[1.0em]
    \multirow{3}{*}[0.1em]{Pretrained GPT-2}
    &
   The shirt was completely ripped with no stitching, it was a mess with no color or anything else that suggested anything.  This is now in a shop.  - Michael I'm  very, very sorry about this experience and that they will consider me a part of a larger family of customers.
  \\
    \addlinespace[1.0em]
    \multirow{3}{*}[0.1em]{Fine-tuned GPT-2}
    &
    This store is \textbf{disgusting}.  I went in a couple weeks ago to pick up a blouse of mine.  The manager on duty was \textbf{extremely rude} and made me feel like I was interrupting her personal conversation.
    \\
    \addlinespace[1.0em]
    \multirow{3}{*}[0.1em]{mLSTM}
    &
    I will \textbf{never go} back to this place. The staff at this  location are \textbf{the worst} I have ever encountered in a restaurant. The food is \textbf{overpriced} and the servers are so \textbf{rude} to them and they will take your business to a club that is close to me.
\\
    \addlinespace[1.0em]
    \multirow{3}{*}[0.1em]{Sentiment modeling}
    &
    I will \textbf{never go} to this store again.  I would \textbf{never go} back to this store again.  The store is still okay but \textbf{not worth} the price they charge for everything.
    \\
    \bottomrule
    
  \end{tabular}
  }
  \vspace{-5mm}
\end{table*}

\vspace{-7mm}
\subsection{Subjective Evaluation}
\vspace{-5mm}
As shown in Table~\ref{Tab:naturalness}, the non-native English speakers tended to give higher scores for fluency than the native English speakers to the original reviews while the native speakers tended to give higher scores to most cases of fake reviews (5 of 8), but the differences are slight. The fine-tuning improved the fluency compared with that of the reviews generated by the pre-trained GPT-2. This suggests that an attack can be made more effective by simply fine-tuning existing models. For the Amazon dataset, the reviews generated by explicitly modeling the sentiment (sentiment modeling) had the highest overall score, followed by those generated by the fine-tuned GPT-2 model. Interestingly, the scores for all fake review were higher than that for the original review. This observation is similar to that of Yao et al.~\cite{yao2017automated}, who observed that people tended to consider fake reviews highly reliable. This observation does not hold for the Yelp database --- the score for the original reviews is higher than those for the fake ones. Among the fake review generation models, the fine-tuned GPT-2 model had the highest score (3.30).

Table \ref{Tab:judgement} shows the results for judging which of the listed four reviews was the most likely real review. It was surprising to find that it was difficult to identify the real review from the four options. The lowest overall correctness were 25.4\% and 20.8\% and the highest ones were 29.1\% and 34.6\% for the Amazon and Yelp databases, respectively. These results demonstrate that the participants tended to randomly judge which of the listed four reviews was the most likely real review because the rates were close to the chance rate of 25\%.

\begin{table}[t]
    \centering
    \footnotesize
    \caption{Fluency of reviews (in MOS). Bold font indicates highest score.
       }
    \label{Tab:naturalness}
    \scalebox{0.8}{
   \begin{tabular}{p{32mm} lcp{20mm}lcc}
   \toprule
    \multirow{2}{*}{Model} &  & Amazon &  &  & Yelp &  \\ 
     & Native & Non-native & Overall  &  Native & Non-native & Overall \\
    \midrule
    Original review           & $2.85$   &  $3.09$ & $2.95$   &  
    $\mathbf{3.43} $ & $\mathbf{3.56} $   &  $\mathbf{3.49}$  \\
    Pretrained GPT-2          & $2.93$   &  $3.16$ & $3.06$   &  
    $2.68 $ & $2.72 $   &  $2.70$  \\
    Fine-tuned GPT-2          & $3.24$   &  $3.22$ & $3.23$   &  
    $3.35$ & $3.25 $   &  $3.30$  \\
    mLSTM         & $3.06$   &  $\mathbf{3.37}$ & $3.21$   &
    $3.12 $ & $2.96 $   &  $3.04 $  \\
    Sentiment modeling         & $\mathbf{3.61}$   &  $3.35$ & $\mathbf{3.47}$   &  $2.90$ & $2.86$ & $2.88$ \\
    \bottomrule
    \end{tabular}
    }
    \vspace{-3mm}
\end{table}
\begin{table}[t]
    \centering
    \footnotesize
    \caption{Correctness (in \%) for judging which of four reviews was the most likely real review. Bold font indicates worst case. 
     }
    \label{Tab:judgement}
    \scalebox{0.8}{
   \begin{tabular}{p{32mm} lcp{20mm}lcc}
   \toprule
    \multirow{2}{*}{Model} &  & Amazon &  &  & Yelp &  \\ 
     & Native & Non-native & Overall  &  Native & Non-native & Overall \\
    \midrule
    Pretrained GPT-2          & $30.5$   &  $27.9$ & $29.1$   &  
    $\mathbf{20.0}$ & $\mathbf{21.8}$  &  $\mathbf{20.8}$ \\
    Fine-tuned GPT-2          & $28.6$   &  $\mathbf{23.6}$ & $25.9$   &  
    $30.0$ & $26.9$  &  $28.3$ \\
    mLSTM         & $\mathbf{22.0}$   &  $28.4$ & $\mathbf{25.4}$   &  $32.8$ & $36.5$  &  $34.6$ \\
    Sentiment modeling         & $23.8$   &  $34.4$ & $29.1$   & $22.4$ & $31.3$ & $26.7$ \\
    \bottomrule
    \end{tabular}
    }
    \vspace{-4mm}
\end{table}

\vspace{-7mm}
\subsection{Automatic Fake Review Detection}
\vspace{-5mm}
We evaluated the three automatic detection methods using 80 real reviews and 160 fake reviews per database. We used another set consisting of 120 real reviews and 240 fake reviews per database for training the regression models used as fusion functions.
Table~\ref{Tab:detection} shows detection result. For a single detector, the lowest EER of 23.5\% was achieved by the GPT-2PD. When Grover and GTLR were fused, the EER was greatly reduced compared to individual detectors. When Grover or GTLR was combined with GPT-2D, the EER on Amazon dataset was increased while the EER on Yelp dataset was reduced. The lowest EER of 22.5\% was achieved by fusing the three detectors or fusing GTLR and GPT-2PD. 
The accuracy of the automatic detection is higher than the chance level and it means that there are some traces found for identifying the fake reviews. However, the overall EERs are very high and we see that it is not straightforward to precisely detect the fake reviews generated by the proposed low-skilled method.

\begin{table*}[t]
    \centering
    \caption{Equal error rate in distinguishing between fake and real reviews. GPT-2PD is the pre-trained detector recently released by OpenAI. ``+'' indicates score fusion.}
    \label{Tab:detection}
    \scalebox{0.8}{
   \begin{tabular}{lccc}
   \toprule
    \textbf{Detector}   & \textbf{Amazon}  &     \textbf{Yelp} &    \textbf{Overall} \\ 
    \midrule
    Grover  &  $43.6\%$  &  $36.9\%$ &  $40.7\%$\\ 
    GTLR & $40.9\%$  & $35.9\%$ & $38.5\%$\\ 
    GPT-2PD  &  $\mathbf{20.9\%}$ & $25.8\%$ & $23.5\%$\\ 
    \midrule
    Grover + GTLR     & $35.3\%$  & $34.6\%$ & $34.9\%$\\
    Grover + GPT-2PD  & $24.9\%$  & $22.2\%$ & $23.4\%$\\ 
    GTLR + GPT-2PD    & $25.0\%$  & $\mathbf{19.6\%}$ &  $\mathbf{22.5\%}$\\
    \shortstack{Grover + GTLR + GPT-2PD } & $25.0\%$  & $\mathbf{19.6\%}$ & $\mathbf{22.5\%}$\\
    \hline
    \end{tabular}
    }
    \vspace{-3mm}
\end{table*}

\vspace{-7mm}
\section{Conclusion}
\vspace{-5mm}
We proposed a low-skilled and sentiment-preservable
fake review generation method. It fine-tunes GPT-2 model to generate a large number of reviews based on a review with the desired sentiment taken from the website to be attacked. Then it uses the BERT text classifier to filter out the ones with undesired sentiments. Since there is no post-processing or word modification, the generated reviews may be as fluent as the samples used for language model training. Subjective evaluation of review fluency by 80 participants  produced a mean opinion score of 3.23 (scale of $1-5$) for fake reviews based on Amazon real reviews and 3.30 for fake reviews based on Yelp real reviews. The values for the real reviews were 2.95 and 3.49, respectively. This means that the generated reviews had the same fluency as the reviews written by a person. Subjective judgment of which of four reviews (one real review and three fake reviews in random order) was the most likely real review produced correctness between 20.8\% and 34.6\%. This is roughly equivalent to random selection. Application of three countermeasures to the detection of fake reviews, Grover, GLTR, and GPT-2 detector, demonstrated a detection equal error rate of 22.5\%. Although the error rate of automatic detection methods is smaller than chance and has better performance than humans, these methods are still far from perfect, meaning further improvements are needed.

We plan to investigate ways to further preserve both sentiment and context information by using cold fusion~\cite{sriram2017cold} or simple fusion~\cite{stahlberg2018simple}. Since the generated reviews is the most probable sequence, they lack diversity and the corresponding distribution area may be already covered by the countermeasures. 
This would further increase detection errors.
We also plan to develop a countermeasure for detecting these generated reviews.

\vspace{2mm}
\noindent
\small{\textbf{Acknowledgments:} This research was carried out when the first and second authors were at the National Institute of Informatics (NII) of Japan in 2018 and 2019 as part of the NII International Internship Program. This work was partially supported by a JST CREST Grant (JPMJCR18A6) (VoicePersonae Project), Japan, and by MEXT KAKENHI Grants (16H06302, 17H04687, 18H04120, 18H04112, 18KT0051), Japan.}

\vspace{-7mm}
%
%
 \bibliographystyle{IEEEtran}
\bibliography{reference}

\end{document}